\documentclass[sigconf]{acmart}
\usepackage{enumitem}
\usepackage{mathtools}
\usepackage{natbib}
\usepackage{float}
\usepackage{bm}
\usepackage[switch]{lineno}
\usepackage{amsfonts}
\usepackage{xspace}
\usepackage{booktabs}
\usepackage{url}

\usepackage[ruled,vlined]{algorithm2e}
\usepackage{amsmath,bm}
\usepackage{dsfont}
\usepackage{multirow}
\usepackage[ruled,vlined]{algorithm2e}
\usepackage[autostyle, english = american]{csquotes}
\usepackage{xcolor}
\usepackage{subcaption} 
\theoremstyle{acmplain}

\newtheorem{problem}{Problem}
\newcommand{\ie}{\emph{i.e.}}

\newcommand{\baselineNotation}{}

\newcommand{\LSTM}{\baselineNotation{LSTM}\xspace}
\newcommand{\RETAIN}{\baselineNotation{RETAIN}\xspace}
\newcommand{\Dipole}{\baselineNotation{Dipole}\xspace}
\newcommand{\MiME}{\baselineNotation{MiME}\xspace}
\newcommand{\DMEGP}{\baselineNotation{DME-GP}\xspace}
\newcommand{\INPREM}{\baselineNotation{INPREM}\xspace}
\newcommand{\DE}{\baselineNotation{DeepEnsemble}\xspace}
\newcommand{\MC}{\baselineNotation{MC-Dropout}\xspace}

\newcommand{\ours}{\texttt{UNITE}\xspace}
\AtBeginDocument{%
  \providecommand\BibTeX{{%
    \normalfont B\kern-0.5em{\scshape i\kern-0.25em b}\kern-0.8em\TeX}}}

\setcopyright{iw3c2w3}
\copyrightyear{2021}
\acmYear{2021}
\acmDOI{10.1145/3442381.3450087} 
\acmConference[WWW '21]{Proceedings of the Web Conference 2021}{April 19--23, 2021}{Ljubljana, Slovenia}
\acmBooktitle{Proceedings of the Web Conference 2021 (WWW '21), April 19--23, 2021,
Ljubljana, Slovenia}
\acmPrice{}
\acmISBN{978-1-4503-8312-7/21/04}
\begin{document}

%%
%% The "title" command has an optional parameter,
%% allowing the author to define a "short title" to be used in page headers.
\title{\ours: Uncertainty-based Health Risk Prediction  \\ Leveraging Multi-sourced Data}

%%
%% The "author" command and its associated commands are used to define
%% the authors and their affiliations.
%% Of note is the shared affiliation of the first two authors, and the
%% "authornote" and "authornotemark" commands
%% used to denote shared contribution to the research.
\author{Chacha Chen}
\affiliation{%
  \institution{Penn State}
}
\email{chachachen@psu.edu}

 \author{Junjie Liang }
 \affiliation{%
  \institution{Penn State}
}
\email{jul672@ist.psu.edu }
 \author{Fenglong Ma}
 \affiliation{%
  \institution{Penn State}
}
\email{fenglong@psu.edu}
 \author{Lucas M. Glass }
 \affiliation{%
  \institution{IQVIA and Temple University}
}
\email{Lucas.Glass@iqvia.com}
 \author{Jimeng Sun }
 \affiliation{%
  \institution{UIUC}
}
\email{jimeng.sun@gmail.com}
 \author{Cao Xiao }
 \affiliation{%
  \institution{IQVIA}
}
  \email{danicaxiao@gmail.com}

%%
%% By default, the full list of authors will be used in the page
%% headers. Often, this list is too long, and will overlap
%% other information printed in the page headers. This command allows
%% the author to define a more concise list
%% of authors' names for this purpose.
\renewcommand{\shortauthors}{Chen and Liang, et al.}

%%
%% The abstract is a short summary of the work to be presented in the
%% article.
\begin{abstract}
Successful health risk prediction demands {\bf accuracy} and {\bf reliability} of the model. 
Existing predictive models mainly depend on mining electronic health records (EHR) with advanced deep learning techniques to improve {\bf model accuracy}. However, they all ignore the importance of publicly available online health data, especially socioeconomic status, environmental factors, and detailed demographic information for each location, which are all strong predictive signals and can definitely augment precision medicine.
To achieve {\bf model reliability}, the model needs to provide accurate prediction and uncertainty score of the prediction.
However, existing uncertainty estimation approaches often failed in handling high-dimensional data, which are present in multi-sourced data.

To fill the gap, we propose \underline{UN}certa\underline{I}n\underline{T}y-based h\underline{E}alth risk prediction (\ours) model. Building upon an adaptive multimodal deep kernel and a stochastic variational inference module, \ours provides accurate disease risk prediction and uncertainty estimation
leveraging multi-sourced health data 
%\cx{including location based public health data collected from the web} on the web including location, socioeconomic and environmental factors. 
including EHR data, patient demographics, and  public health data collected from the web.
We evaluate \ours on real-world disease risk prediction tasks: nonalcoholic fatty liver disease (NASH) and Alzheimer's disease (AD). \ours achieves up to 0.841 in F1 score for AD detection, up to 0.609 in PR-AUC for NASH detection, and outperforms various state-of-the-art baselines by up to $19\%$ over the best baseline. We also show  \ours can model meaningful uncertainties and can provide evidence-based clinical support by clustering similar patients.

\end{abstract}

%%
%% The code below is generated by the tool at http://dl.acm.org/ccs.cfm.
%% Please copy and paste the code instead of the example below.
%%


%%
%% Keywords. The author(s) should pick words that accurately describe
%% the work being presented. Separate the keywords with commas.

%% A "teaser" image appears between the author and affiliation
%% information and the body of the document, and typically spans the
%% page.

%%
%% This command processes the author and affiliation and title
%% information and builds the first part of the formatted document.
\maketitle

\section{Introduction}\label{intro}

In recent years, both patient electronic health records (EHR) and public health data are increasingly accumulated and available for researchers, which significantly advances the field of health informatics, especially for the health risk prediction task. Existing health risk prediction models mainly depend on mining EHR data with advanced deep learning (DL) techniques, which often capture disease patterns based on patients' personal health trajectory ~\cite{choi2016retain,baytas2017patient,ma2017dipole,choi2018mime,ma2020concare}  and output a probability score that indicates the likelihood of a patient who will develop a target disease. 

\smallskip
\noindent\textbf{Multi-sourced Health Data Integration}. However, existing DL models all ignore the importance of online public health data especially socioeconomic status, environmental factors and detailed demographic information for each location. In fact, many of these public health data are strong predictive signals and can definitely augment precision medicine such as patient health risk prediction. For example,  studies have shown that environmental factors become important risk factors to the onset and progression for diseases like Alzheimer's disease (AD)~\cite{killin2016} and nonalcoholic steatohepatitis (NASH)~\cite{Kneeman12}. Thus, \emph{how to integrate these data from difference sources to enhance the prediction performance is a new challenge for risk prediction task}.

\smallskip
\noindent\textbf{Model Uncertainty}.
Since existing DL models only output a probability score, which are known to be prone to overconfidence and are difficult to produce effective patient-level uncertainty scores \cite{li2020deep}. This affects the model reliability especially in the clinical context. Since even a model has  good average performance for a patient population, there still can be out-of-distribution (OOD) patients whose predictions are quite uncertain but often difficult to detect. 

\smallskip
To provide uncertainty estimation, the probabilistic deep learning models including Bayesian neural networks (BNN)~\cite{Zhang2019AdvancesIV,tran2019bayesian}  and sparse Gaussian process (GP)~\cite{liu2018gaussian} has drawn recent interest. BNN models such as~\cite{10.1145/3394486.3403087} learn a distribution over model parameter weights and use the uncertainty of the parameter weights to estimate the uncertainty of predictions. However, this approach usually requires a compromise between model complexity and expressiveness of variational distributions. Recently, there are also some attempts \cite{lakshminarayanan2017simple,gal2016dropout} in modifying DNNs to approximate Bayesian, but they are not as flexible as non-parametric models such as GP models. Compared with BNN models, the GP models such as ~\cite{Chung2020,futoma2017learning,futoma2017improved,futoma2018learning} are more expressive, but at the computational expense of the need to store and process the data points for the covariance matrix. These disadvantages make existing GP-based approaches unable to model high-dimensional EHR data with reasonable efficiency and expressiveness. Therefore, \emph{it is essential and necessary to design an effective health risk prediction model to integrate multi-sourced health data and handle model uncertainty issue simultaneously}.

\smallskip
\noindent\textbf{Our Approach}. To tackle all the aforementioned challenges, we propose \ours, an uncertainty-based risk prediction model. Built upon an adaptive multimodal deep kernel and a stochastic variational inference module, \ours provides accurate disease risk prediction and uncertainty estimation
leveraging multi-sourced health data including including public health data and patient EHR. \ours is enabled by the following technical contributions.

\begin{enumerate}[leftmargin=*]
    \item \textbf{A stochastic variational inference module to provide accurate prediction and patient-level uncertainty score}. To provide informative predictions based on global correlations, we formulate a population-level GP classification task. We introduce non-Gaussian likelihood, stochastic variational inference (SVI), and inducing points to enhance computational tractability.
    Moreover, \ours is able to provide meaningful patient-level uncertainty scores, which is a key step in applying these models in clinical decision support applications. 
    \item \textbf{An adaptive deep kernel to integrate  location-based public health risk factors}. The deep kernel module first applies to different data modalities, then uses a weight-based fusion network to learn the shared representation across all data modalities. The deep kernel module can effectively extract latent feature representations from multiple data sources.
    
    \item \textbf{Comprehensive evaluation and benchmarking on health risk prediction}
    We compare \ours with 8 baselines (4 deep learning and 4 probabilistic deep learning methods) and 2 variants of \ours on two disease prediction tasks for 
    Nonalcoholic fatty liver disease (NASH) and Alzheimer. We utilize two large real-world longitudinal patient datasets and their related location based public health data in the experiment. Overall, \ours significantly outperformed all the other baselines with .609 PRAUC for NASH and 0.927 in Cohen's Kappa for AD. We also show the learned uncertainty scores could potentially be used to identify truly uncertain patients that are easily misdiagnosed, thus improving the overall prediction.
\end{enumerate}

\section{Related Work}
\label{related}

\noindent\textbf{Deep learning models for health risk prediction} mainly extract temporal patterns or graph structures from patients' historical diagnosis and medication records in their EHR data. For example, ~\cite{LiptonKEW15,choi2016doctor,baytas2017patient,suo2017multi,zhang2019interpretable} leveraged  recurrent neural networks and its variant to better capture temporal health patterns. \cite{choi2016retain,ma2017dipole} used bidirectional RNN and attention to capture the relationships of different visits for the prediction. \cite{choi2017gram,ma2018kame,zhang2019knowrisk} exploited graph-based attention mechanism to learn general knowledge representations for health risk prediction.
\cite{choi2018mime} and \cite{choi2020learning} learned and exploited the graphical structure inherent in medical codes. In addition, \cite{8594952,gao2019camp,ma2020concare} also considered patient demographics information the their models.
However, in the real world, there are other factors, such as location-based health factors,  also affect the health risks. 

In this work, we aim to utilize data from multiple sources and build \ours as a multimodal learning framework. Moreover, our work innovatively provides prediction along with uncertainty scores in an effort of promoting clinical applications.\\

\noindent\textbf{Probabilistic deep learning models for predictive health} include Bayesian neural networks (BNN)~\cite{Zhang2019AdvancesIV} and sparse Gaussian process (GP)~\cite{liu2018gaussian} has drawn recent interest.  
For example, \cite{10.1145/3394486.3403087} proposed to use Bayesian neural networks (BNN) to capture model uncertainty. It learns a distribution over model weights and pass the uncertainty of the weights to eventually estimate the uncertainty in the predictions. However, this approach usually requires a compromise between model complexity and expressiveness. Also it could not learn population-level correlation among patients.

Compared with BNN models, GP models are more flexible and expressive. Recent attempts include \cite{futoma2017learning,futoma2017improved,futoma2018learning}, which used multi-task GP models for individual sepsis detection. Similarly, \cite{peterson2017personalized} proposed a multiple GPs formulation to diagnose the Alzheimer's disease. \cite{cheng2020sparse} developed a multi-output Gaussian process regression model for hospital patient monitoring. However, since GP models are expensive in storing data points for the covariance matrix, these works either focus on low dimensional numerical features or decompose the task into separate modeling without considering the global effect. Recently,
\cite{Chung2020} proposed to augment GP with a deep mean function for learning the global trend and a deep kernel for individual variability. However, it could not learn population-level correlation among patients. 
There are also attempts \cite{lakshminarayanan2017simple,gal2016dropout,DBLP:journals/corr/abs-1906-03842} in modifying DNNs to approximate Bayesian, but are not as flexible as non-parametric models such as GP.

In this work, we propose a GP-based model that has the following benefits: (1) It inherits the expressive power of GP and is augmented by a multimodal deep kernel that fuses multi-sourced data for improved representation power. (2) It reduces the computation cost by leverage stochastic variational inference with inducing points~\cite{wilson2016deep,wilson2016stochastic} as an approximate  to handle large-scale input data. (3) The covariance matrix provides population-level patient correlations, which could benefit evidence-based medicine.\\

\noindent\textbf{Combining GP and deep learning models} has drawn recent interest. Combined model can benefit from both the flexibility of kernel methods and from the structural properties of deep neural networks.
% \noindent\textbf{Combining GPs and deep learning models}
% There has been a growing interest in combining deep architectures with GP to benefit from both methods. 
\cite{hinton2008using} proposed to use a deep belief networks to learn a good covariance kernel in a semi-supervised learning framework. 
\cite{wilson2011gaussian} developed Gaussian process regression network by replacing all weight connections in a Bayesian neural network with GP. Similarly, \cite{damianou2013deep} replaced the activation function in a Bayesian neural network with GP in a unsupervised learning setting. \cite{calandra2016manifold} combined the GP with a feed forward neural network transformation. However, aforementioned methods are task specific and can not scale beyond a few thousand training points. Alternatively,
\cite{huang2015scalable} transformed the inputs of the base kernels with a deep architecture by applying deep neural networks as the feature-mapping function of GPs. 
More generally, \cite{wilson2015kernel,wilson2015thoughts,wilson2016deep} proposed deep kernel learning by parameterizing the kernels with neural networks to augment the model with the expressiveness power from deep architectures. Moreover, another line of works \cite{damianou2013deep,salimbeni2017doubly} proposed a stacked GP model which has multiple layers of Gaussian processes. 

In this work, we augment the deep kernel learning framework by learning a deep multimodal kernel function parameterized by well-designed deep neural networks.

\section{Method}
\label{method}
\subsection{Problem Formulation}
\label{formulation}
Given longitudinal EHR  $\boldsymbol{E} \coloneqq  \{\boldsymbol{X_i},y_i\}_{i=1}^P$ of $P$ patients, where patient $i$ is represented by a multivariate sequence $\boldsymbol{X_i} \coloneqq\{x_1^{(i)},x_2^{(i)}...x_{T_i}^{(i)}\}$. Here $x_{T_i}^{(i)}$ represents a medical code indicating the diagnosis or medication that patient $i$ receives at time $T_i$. The groundtruth targets are given as $y_i\in\{0,1\}$ where 1 means the patient has the disease and 0 means not. In addition to the EHR dataset, we also have the access to the patient's demographics data $\boldsymbol{D} \coloneqq  \{\boldsymbol{S_i}\}_{i=1}^P$ of $P$ patients. $\boldsymbol{S_i}\coloneqq  \{s_1^{(i)},s_2^{(i)},...,s_N^{(i)}\}$ denotes the demographics of the patient $i$.

% $\boldsymbol{D}\coloneqq  \{s_1,s_2,...,s_P\}_{j=1}^P$,
% where each $s_i$ denotes the demographics feature of the $i$-th patient. Specifically, we have the patient's ID, gender, age and zipcode. 

Moreover, we denote the location-based health risk data (open-sourced on CDC website) as $\boldsymbol{G} \coloneqq  \{\boldsymbol{G_k}\}_{k=1}^L$ of $L$ locations.
$\boldsymbol{G_k} \coloneqq  \{g_1^{(k)},g_2^{(k)},...,g_M^{(k)}\}$, where each $g_i^{(k)}$ denotes one of the location-based factors at location $k$. 

% For the detailed data statistics, please refer to Table~\ref{tab:dataset_des}.

\begin{problem}[Uncertainty-based Health Risk Prediction]
Given the data from multiple modalities $\boldsymbol{E}, \boldsymbol{D}, \boldsymbol{G}$, our objective is to provide  informative health risk prediction. Specifically, for a patient $i$ at location $k$ with EHR input $\{\boldsymbol{X_i}\}$, demographics features $\{\boldsymbol{S_i}\}$, and the location-based factors $\{\boldsymbol{G_k}\}$, the model aims to output a health risk distribution $p(y)$, such that (1) by evaluating the mean of $p(y)$, \ie, $\mathbb{E}(y)$, we can access the predicted health risk that indicates the probability of having the target disease; and (2) by computing the variance of $p(y)$, \ie, $\text{var}(y)$, we can quantify the uncertainty estimations about the model predictions.
\end{problem}

\begin{table}[h!]
\centering
\caption{Notations used in \ours}
\label{tab:notation}
\resizebox{\columnwidth}{!}{
\begin{tabular}{cl}
\toprule
  Symbol        & Definition and description \\
          \midrule
 $\boldsymbol{E}$ &  The EHR dataset    \\
  $\boldsymbol{X_i}$ &  The EHR record of patient $i$    \\
    $x_T^{i}$ &  The medical code of patient $i$ at time $T$   \\
%  $M_s$ & Symptom medical code set\\
%  $M_d$ & Diagnosis medical code set\\
%  $m_i^{t}$ & $i$-th medical code in $t$-th visit\\
%  $X_p$ & The visit record of patient $p$\\
%   $x_p^{t}$ & The $t$-th visit record of patient $p$\\
$\boldsymbol{D}$ &  The demographics dataset    \\
$s_j^{(i)}$ &  The $j-th$ demographics features of patient $i$    \\
  $\boldsymbol{G}$ &  The location-based  public health  dataset    \\
  $g_l^{(k)}$ &  The $l-th$ location-based features at location $k$    \\
 $y\in\{0,1\}$ & Patient's disease label     \\
\bottomrule
\end{tabular}}
\end{table}

 \begin{figure*}[h!]
  \centering
  \includegraphics[width=0.88\linewidth]{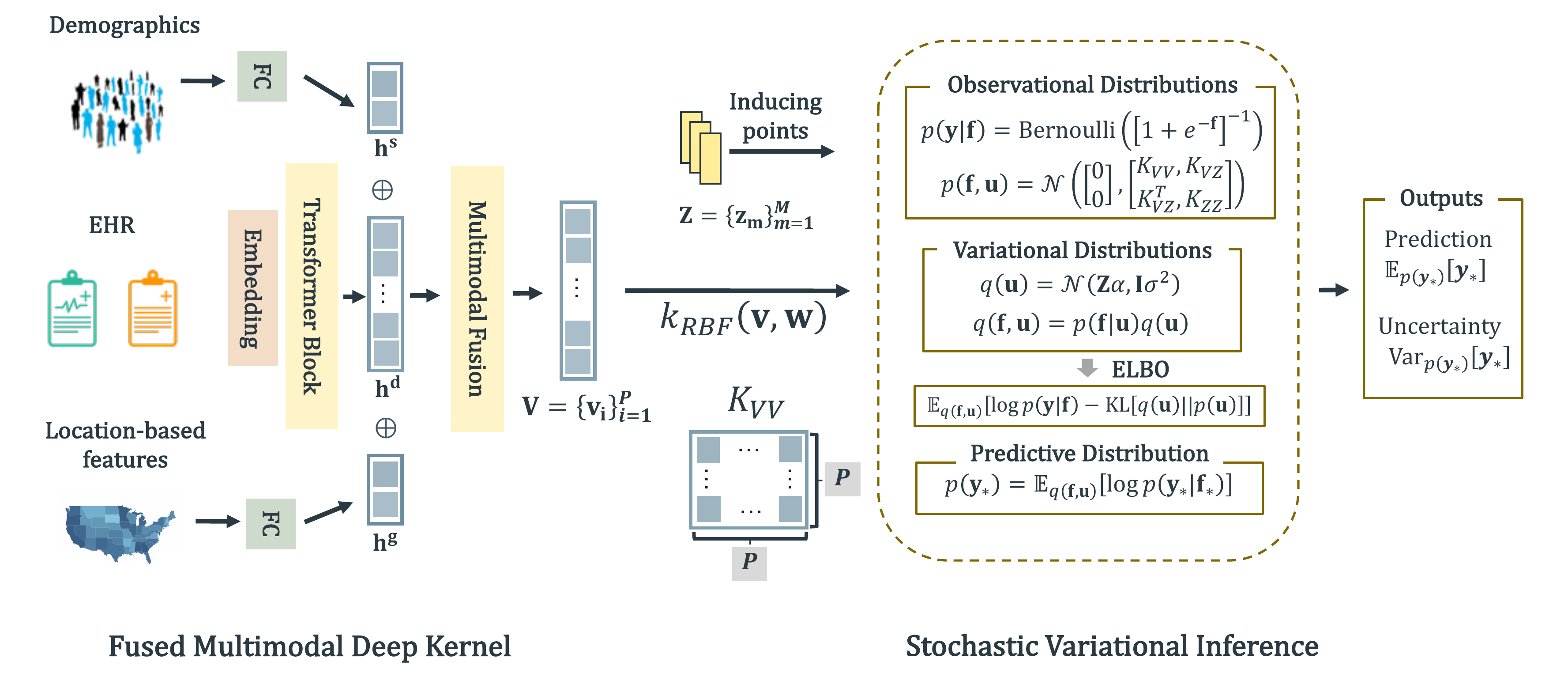} 
%   \vspace{-0.1in}
  \caption{The \ours model comprises of two components: the \textbf{fused multimodal deep kernel module} and \textbf{stochastic variational inference module}. \textbf{(I)} In fused multimodal deep kernel, we integrate data from multiple sources with different embedding mechanism, and then fuse them using a weight-based fusion technique to learn a shared embedding. After that, we warp the embedding networks with an radial basis function (RBF) kernel. \textbf{(II)} The embedding are then fed into a stochastic variational inference (SVI) module for making predictions as well as generating uncertainty scores based on global correlations. We also equip the SVI module with inducing points for efficient training.
} 
  \label{fig:training}
%   \vspace{-0.1in}
\end{figure*}

\subsection{The \ours Model}

 \ours comprises of two modules: the fused multimodal deep kernel module and the stochastic variational inference module. The fused multimodal deep kernel module first integrates data from multiple sources with different embedding mechanism, fuses them using a weight-based fusion technique to learn a shared embedding, and  warps the embedding networks with a radial basis function (RBF) kernel. While stochastic variational inference (SVI) module takes the learned embedding
to make predictions as well as perform uncertainty estimation based on global correlations. It is also equipped with an inducing point for efficient training.

\subsection*{(I) Fused Multimodal Deep Kernel Module}

We leverage deep neural structures to effectively learn the meaningful feature representations from rich data. Specifically, we design different embedding frameworks for various data modalities. Then to learn a shared representation of comprehensive input data, we leverage an adaptive information fusion mechanism. The  feature embedding networks are pre-trained using a linear classifier with cross-entropy loss to capture meaningful latent representations. The final step is to warp the latent vectors with a base RBF kernel.

\noindent\subsubsection{EHR Records Embeddings} The EHR record of a patient includes a sequence of the diagnosis and medical codes for each visit. To learn the sequence embedding, we first tokenize the medical codes into sequences of integers (integers are indexes of tokens in a dictionary) as given by Eq.~\eqref{eq:index}. 
    \begin{equation}
         \boldsymbol{h_0^d} = \mathrm{Tokenizer}(\boldsymbol{X_i})
         \label{eq:index}
    \end{equation}
Then we utilize a multi-layer transformer to learn the latent representations. Transformer, which is a multi-head attention based architecture, is able to learn the spatial relationships in the patient EHR records. In general, the attention mechanism maps a query and a set of key-value pairs to an output. The output is the weighted sum of the values, as the weight computed by a function of the corresponding query key pair. Since the self-attention mechanism contains no recurrence, it is a common practice that we add a position encoding to provide the sequential relationships of the input sequence as well \cite{song2017attend}. 

We build multi-layer transformer blocks to capture the rich information in high-dimensional  data. In detail, each transformer block sequentially applies a multi-head self attention layer, a normalization layer, a feed forward layer, and another normalization layer.  We also add residual connections before the normalization to help the networks train faster.
Specifically, given EHR  record for patient $i$: $\boldsymbol{X_i}\coloneqq[x_1^{(i)},...x_{T_i}^{(i)}]$, the latent embedding is given by Eq.~\eqref{eq:transformer}.
\begin{equation}
     \boldsymbol{h_{n+1}^d} =\mathrm{Transformer}(\boldsymbol{h_{n}^d})
     \label{eq:transformer}
\end{equation}
where $n$ is a hyperparameter indicating the number of transformer blocks applied.

% \noindent   
\smallskip
\subsubsection{Demographics and Location-based Features Embeddings}
    Furthermore, the demographics and location-based features  from different locations are embedded into the same hidden space. For those numerical and categorical data, we leverage multi-layer perceptrons (MLP) to learn the latent representations as given by Eq.~\eqref{eq:mlp}. 
    \begin{equation}
      \boldsymbol{h^s} = \mathrm{MLP}(\boldsymbol{s}) ; \quad
      \boldsymbol{h^g} = \mathrm{MLP}(\boldsymbol{g})
      \label{eq:mlp}
    \end{equation}

\subsubsection{Adaptive Information Fusion}

We demonstrate how to fuse the learned three latent embeddings $\boldsymbol{h^d},\boldsymbol{h^s},\boldsymbol{h^g}$ from different data modalities. To efficiently learn a shared embedding across them, we adopt a parametric-matrix-based fusion technique as given by 
 Eq.~\eqref{eq:v}.
\begin{equation}
    \boldsymbol{v} = \boldsymbol{h^d}\circ\boldsymbol{W_d} \oplus \boldsymbol{h^s}\circ\boldsymbol{W_s} \oplus \boldsymbol{h^g}\circ\boldsymbol{W_g},
    \label{eq:v}
\end{equation}
where $\circ$ is a Hadamard product, i.e., element-wise multiplication, and $\oplus$ is the vector concatenation operator. $\boldsymbol{W_s},\boldsymbol{W_d},\boldsymbol{W_g}$ are the learnable parameters that adjust the weights of the latent factors from different data sources.

Note that our proposed data fusion mechanism retains the flexibility to integrate any additional data modality by concatenating the modality embedding transformed by the corresponding weight matrix.

\subsubsection{Pre-training Initialization}
We pre-train the fused multimodal embedding networks for better initialization, following the implementation in \cite{wilson2016stochastic}. We first apply a fully connected layer to the embedding outputs as a classifier. Then we pre-train the whole networks using Adam optimizer with the cross entropy loss objective in a supervised manner. In practice, by pre-trained initialization, we give the model a better prior to help the model converge fast.

\subsubsection{RBF Kernel Warping} Using the individual level latent vectors obtained from Eq.~\eqref{eq:v}, we can learn the population level covariance matrix among all the patients. Hence, we warp the fused latent vector outputs with a base kernel. In this work, we adopt the RBF kernel as it is widely-used in existing deep kernel method  \cite{wilson2016deep}. Given a pair of latent vectors $\boldsymbol{v,w}\in \boldsymbol{V}$, the RBF kernel function is defined as in Eq.~\eqref{eq:rbf}.
\begin{equation}
\label{eq:rbf}
k_{RBF}(\boldsymbol{v,w}|\gamma) = \exp\left(-\frac{1}{2}\sum_{k=1}^D \left(\frac{v_{k}-w_{k}}{l_k}\right)^2\right)
\end{equation}
where $\boldsymbol{l}=\{l_k\}_{k=1}^D$ is the vector of length scale parameters and $D$ is the size of latent vector. For the rest of the paper, we denote the kernel matrix computed on $\boldsymbol{V}$ as $K_{VV}$.

\subsection*{(II) Stochastic Variational Inference Module}
Exact inference and learning in Gaussian process model with a non-Gaussian likelihood is not analytically tractable. Motivated by \cite{wilson2016deep}, we leverage stochastic variational inference with inducing points as an approximate technique to deal with large-scale input data.

Specifically, given the fused vector for patient $i$ as $\boldsymbol{V} = \{\boldsymbol{v_i}\}_{i=1}^P$, we use Bernoulli likelihood to tackle the classification problem as in Eq.~\eqref{eq:svi1}.
\begin{equation}
     (\boldsymbol{y}|\boldsymbol{f}) \sim \text{Bernoulli}(\frac{1}{1+e^{-\boldsymbol{f}}}); \quad
     \boldsymbol{f}(\boldsymbol{V}) \sim \mathcal{N}(0, K_{VV})
     \label{eq:svi1}
\end{equation}
where $\boldsymbol{f}$ is generally known as the signal function that follows a Gaussian process prior. $K_{VV}$ is the covariance matrix (a.k.a. kernel) as defined in Eq.~\eqref{eq:rbf}. 

Exact model inference for such Gaussian process Classification is intractable since the integral between the Gaussian process prior and Bernoulli distribution cannot be solved analytically. Motivated by \cite{wilson2016deep}, we design to solve our model using inducing points and stochastic variational inference. 

Let $\boldsymbol{Z}=\{\boldsymbol{z_m}\}_{m=1}^M$ be the collection of inducing points and $\boldsymbol{u}$ their corresponding signal. We then can express the joint signal distribution as given by Eq.~\eqref{eq:f_given_uxz}.
\begin{equation}
\label{eq:f_given_uxz}
    (\boldsymbol{f,u}) \sim \mathcal{N}\left(
\begin{bmatrix}
    0 \\
    0
\end{bmatrix},\begin{bmatrix}
    K_{VV} & K_{VZ} \\
    K_{VZ}^\top & K_{ZZ}
\end{bmatrix}
\right)
\end{equation}
Therefore, the signal distribution $\boldsymbol{f}$ conditioned on the inducing points is given by Eq.~\eqref{eq:f_given_u}.
\begin{equation}
\label{eq:f_given_u}
    (\boldsymbol{f}|\boldsymbol{u}) \sim \mathcal{N}(K_{VZ}K^{-1}_{ZZ}\boldsymbol{u}, K_{VV} - K_{VZ}K^{-1}_{ZZ}K_{VZ}^\top)
\end{equation}
Let $\gamma=\{\boldsymbol{l},\boldsymbol{Z}\}$ be the model parameters. We seek to learn the parameters by optimizing the marginal likelihood $p(\boldsymbol{y})$. By assuming a variational posterior over the joint signals $q(\boldsymbol{f},\boldsymbol{u})=q(\boldsymbol{u})p(\boldsymbol{f}|\boldsymbol{u})$ and following the derivation in \cite{wilson2016deep}, we obtain the empirical lower bound (ELBO) as in Eq.~\eqref{eq:elbo}.
\begin{equation}
\label{eq:elbo}
    \log p(\boldsymbol{y})\geq \mathbb{E}_{q(\boldsymbol{u,f})}[\log p(\boldsymbol{y|f})-KL[q(\boldsymbol{u})||p(\boldsymbol{u})]]
\end{equation} 

Eq.~\eqref{eq:elbo} can be optimized using any stochastic gradient methods since the likelihood function $p(\boldsymbol{y|f})$ can be factorized over the data instance, \ie,   $p(\boldsymbol{y|f})=\prod_{i=1}^P p(y_i|f_i)$. 
Following the mean-field assumption, we specify the variational signal distribution by $q_{\boldsymbol{\theta}}(\boldsymbol{u})=\mathcal{N}(\boldsymbol{Z}\boldsymbol{\alpha}, \boldsymbol{I}\sigma^2)$, where $\boldsymbol{\theta}=\{\boldsymbol{\alpha},\sigma^2\}$. Optimizing the ELBO in Eq.~\eqref{eq:elbo} will require using Monte Carlo estimation on the first term $\mathbb{E}_{q(\boldsymbol{u,f})}[\log p(\boldsymbol{y|f})]$. In practice, we could first draw samples $\boldsymbol{u} \sim q_{\boldsymbol{\theta}}(\boldsymbol{u})$ and then draw $\boldsymbol{f}$ from $(\boldsymbol{f|u}) \sim p(\boldsymbol{f|u})$ (Eq.~\ref{eq:f_given_u}). With multiple samples of $\boldsymbol{f}=\{\boldsymbol{f}^{(1)},\cdots,\boldsymbol{f}^{(B)}\}$, the first term in Eq.~\eqref{eq:elbo} can be estimated using Eq.~\eqref{eq:monte_carlo}.
\begin{equation}
\label{eq:monte_carlo}
\mathbb{E}_{q(\boldsymbol{u,f})}[\log p(\boldsymbol{y|f})] \approx \frac{1}{B}\sum_{j=1}^B\log p(\boldsymbol{y}|\boldsymbol{f}^{(j)})
\end{equation}

Since the KL divergence term in Eq.\eqref{eq:elbo} is defined on two Gaussian distributions, the quantity can be computed analytically. With the re-parameterization trick for sampling, we are able to \textit{end-to-end} train all the parameters $\gamma,\theta$ along with the parameters associated to the deep kernel.

% \subsection*{(III) Output Prediction and Confidence} 
\subsubsection{Prediction and Uncertainty Estimation}
Given the latent vectors $\boldsymbol{V}_*$ for the test data, the goal of prediction is to find the predictive distribution $p(\boldsymbol{y}_*)$, as given by Eq.~\eqref{eq:predict}.
\begin{align}\nonumber
    p(\boldsymbol{y}_*) &= \int p(\boldsymbol{y}_*,\boldsymbol{f}_*,\boldsymbol{u}) d\boldsymbol{f}_*d\boldsymbol{u} \\
    \label{eq:predict}
    &\approx \mathbb{E}_{q(\boldsymbol{u,f_*})}[p(\boldsymbol{y}_*|\boldsymbol{f_*})] 
\end{align}
The predictive distribution in Eq.~\eqref{eq:predict} is also intractable, we again use Monte Carlo sampling to estimate the predictive mean and variance on each input latent vector. The procedure is equivalent to following Eq.~\eqref{eq:f_given_u} and \eqref{eq:monte_carlo} except that we substitute the latent vectors $\boldsymbol{V}$ with the $\boldsymbol{V}_*$.

With the prediction distribution $ p(\boldsymbol{y}_*)$, we derive the prediction label and the corresponding uncertainty score as 
given by Eq.~\eqref{eq:output}.
\begin{equation}
   \text{Prediction: } \mathbb{E}_{p(\boldsymbol{y}_*)}[\boldsymbol{y}_*]; \quad
  \text{Uncertainty: } \mathrm{Var}_{p(\boldsymbol{y}_*)}[\boldsymbol{y}_*]
  \label{eq:output}
\end{equation}

With the pre-training initializaiton of the embedding networks, we jointly optimize all the hyperparameters, including deep kernel weights and the variational parameters, by backpropagation using Adam optimizer. 
The pseudocode of \ours  is provided in Algorithm~\ref{alg:unite}.
\begin{algorithm}[htbp]
\SetKwInOut{Output}{Output}
\SetAlgoLined
\KwIn{Training set: $\boldsymbol{E} \coloneqq  \{\boldsymbol{X_i},y_i\}_{i=1}^P$,
$\boldsymbol{D} \coloneqq  \{\boldsymbol{S_i}\}_{i=1}^P$, $\boldsymbol{G} \coloneqq  \{\boldsymbol{G_k}\}_{k=1}^L$, $\{\gamma,\theta, w\}$}
% \KwResult{}
\textbf{Deep kernel pre-training initialization}\\
\For{$i\leftarrow 1$ \KwTo \textit{\# pre-train epoch}}{
Sample a batch set of patients $\mathcal{B} \subset\{\boldsymbol{E},\boldsymbol{D},\boldsymbol{G}\} $\\
Optimize the deep kernel parameters $w$ along with a fully connected layer classifier by minimizing the cross entropy loss}
\textbf{Training with stochastic variational inference}\\
 \While{the \textbf{ELBO} not converged}{
  Sample a batch set of patients $\mathcal{B} \subset\{\boldsymbol{E},\boldsymbol{D},\boldsymbol{G}\} $\\
  Draw $B$ samples of $\boldsymbol{u}$ from $q(\boldsymbol{u})$\\
  \For{each patient $i\in\mathcal{B}$}{
    Draw $B$ samples of $\boldsymbol{f}_i$ from $p(\boldsymbol{f|u})$ \\
    Estimate $\textbf{ELBO}_i$ using Eq.~\eqref{eq:elbo} with the Monte Carlo samples $(\boldsymbol{u,f}_i)$\\
  }
  $\textbf{ELBO}=\sum_{i\in\mathcal{B}} \textbf{ELBO}_i$ \\
  Perform stochastic gradient ascent to update $\{\gamma,\theta, w\}$ based on $\textbf{ELBO}$  \tcp*[r]{End-to-end training while fine-tuning $w$.}
 }
 \Output{Model parameters $\{\gamma^*,\theta^*, w^*\}$}
 \caption{The \ours Method \label{alg:unite}  }
\end{algorithm}

\section{Experiment}
\label{experiment}

\subsection{Datasets}\label{sec:dataset}
\textbf{Patient EHR, demographics}: We extract two patient datasets from a large longitudinal claims database\footnote{Dataset can be accessed upon request.}, which include patients’ EHR records as well as patients' demographics information including age, gender, and their geological locations (zip code), ranging from 2011 to 2020. The data statistics can be found in Table~\ref{tab:ehr_statistics}. Specifically, we conduct classification tasks for the following two diseases, i.e. Nonalcoholic fatty liver disease (NASH) and Alzheimer's disease (AD). NASH is a liver disease that occurs in people who drink little or no alcohol. AD is a progressive disease that slowly destroys memory and thinking skills. 

\noindent
\textbf{Public Health Data}\footnote{\url{https://github.com/Chacha-Chen/UNITE/tree/master/CDC-data}}: There is growing evidence that environmental factors and other location-based factors become important risk factors to the onset and progression of AD~\cite{killin2016} and NASH~\cite{Kneeman12}. We collect the following location-based public health statistics, including infant mortality rate, heart disease death rate, stroke death rate, suicide death rate, homicide death rate, drug poisoning death rate, HIV diagnosis rate, Hepatitis B cases, Hepatitis C cases, CLABSI-Standardized infection ratio, adult obesity, youth obesity, adult smoking, youth smoking, adult physical activity, youth physical activity, adult and youth nutrition, colorectal cancer screening rate, influenza vaccination coverage, and child vaccination coverage. 

We integrate patient data with the environmental and location-based features using  patients' zip codes. Patients with the same zip code share the same location-based features $\boldsymbol{G}$. We merge the demographics $\boldsymbol{D}$ based on the unique patient ID.

\begin{table}[htbp]
\caption{Data statistics of patient EHR.} 
% \vspace{-0.1in}
\label{tab:ehr_statistics}
\centering 
\begin{tabular}{lcc}
\toprule
   Dataset       & NASH      &  Alzheimer        \\ 
    \midrule
% Prevalance & 18.75\% &        .                   \\ 
Positive                 & 9640      &   29031      \\  
Negative                 & 41777     &   43424       \\
Ave. \#visit             & 10.34     &    6.59     \\
Max.  \#visit            & 105       &    99        \\
Min.  \#visit            & 2         &     2        \\
Ave. \#medical codes     & 782.17    &    3546.03   \\
\bottomrule
\end{tabular} 
\end{table}

% \paragraph{Other Location-based  data}
% \begin{itemize}
%     \item Income rate 
%     \item Ethnic groups
%     \item Barriers and costs of health care
% \end{itemize}

% \href{https://dhds.cdc.gov/LP?CategoryId=BARRIER&IndicatorId=CBARRIER&ShowFootnotes=true&View=Map&yearId=YR3&stratCatId1=DISSTAT&stratId1=DISABL&stratCatId2=&stratId2=&responseId=YESNO01&dataValueTypeId=AGEADJPREV&MapClassifierId=quantile&MapClassifierCount=5}{\color{blue}CDC website}

\subsection{Baselines and Variants}
We  compare \ours with the following baseline models. For fair comparison, we use all datasets $\{\boldsymbol{E}, \boldsymbol{S}, \boldsymbol{G}\}$ for all the models. For baseline methods, we concatenate the additional features (demographics and location-based features) with the EHR sequence embeddings to further perform the classification. For our proposed \ours model, \textbf{we provide the code in the github \footnote{\url{https://github.com/Chacha-Chen/UNITE}}}. 
% For baseline models, we mainly collect the model code from the authors' website. For more implementation details, please refer to Section~\ref{sec:imple}.
% \cx{Below for each baseline model, be more concrete how we implement these models in our experiments. }

% We  defer the \textbf{code source}, \textbf{implementation details} and \textbf{model setup} to the appendix.
\subsubsection{(a) Deep Learning Models}

\begin{itemize}[leftmargin=*]
    \item \underline{\LSTM} \cite{LiptonKEW15}: the first study of applying LSTM in learning to diagnose. LSTM learns to recognize the patterns in multivariate clinical time series. We feed the EHR sequences into LSTM model, and then concatenate with demographics and location-based features to produce the final prediction.
    \item \underline{\RETAIN} \cite{choi2016retain}: a two-level attention-based neural model for disease risk detection. \RETAIN detects influential past visits and significant clinical variables within those visits such as key diagnoses. It receives the EHR data in a reverse time to mimic the practical clinical course.
    \item \underline{\Dipole} \cite{ma2017dipole} employs a bidirectional RNN to embed the EHR data and applies three attention mechanisms to measure the inter-visit relationships by calculating the weights for previous visits.
    \item  \underline{\MiME} \cite{choi2018mime} leverages the internal multilevel structure of EHR for improving risk prediction. \MiME
    transform the EHR into multilevel embeddings to accurately capture the different patterns of patients.
    \end{itemize}
     
\subsubsection{(b) Deep Learning Models with Uncertainty Estimations}
    \begin{itemize}[leftmargin=*]
    \item \underline{\MC} \cite{gal2016dropout} uses dropout NNs to approximate Bayesian inference in GP. Since \MC is flexible and can be combined with any deep learning models, we build it upon the best deep learning baseline model \Dipole. The dropout rate is set as 0.1 after grid search.
    \item \underline{\DE} \cite{lakshminarayanan2017simple} models the predictive distribution with uncertainty estimation using a proper scoring rule (negative log likelihood) as training criteria and training ensembles of NNs. We build it upon the best deep learning baseline \Dipole. The number of ensemble models is chosen as 10.
    \item \underline{\DMEGP} \cite{Chung2020} leverages a deep mean function learning global effect and a deep kernel function learning the individual trend. We use RNN embedding for deep mean and deep kernel to achieve the best performance on risk prediction. 
    \item \underline{INPREM}~\cite{10.1145/3394486.3403087} captures predictive uncertainty leveraging Monte Carlo sampling and designs an efficient embedding network using attention-based encoding with a linear model for interpretability. 
\end{itemize}

\subsubsection{(c) Variants of \ours}

In addition, we consider two variants of the proposed model. (1) \underline{\texttt{UNITE}$_{d}$}: with a fully-connected layer as the classifier, and (2) \underline{\texttt{UNITE}$_{b}$}: dropout with Monte Carlo sampling for uncertainty estimations.\\

\subsection{Implementation Details}\label{sec:imple}
For each of the AD and NASH detection tasks, we randomly split the dataset into training, validation, and testing sets in a 64:16:20 ratio. The dimension of medical code embedding is set as $128$ for all models. The vocabulary size are  26706 for NASH detection and 31467 for AD detection. The sequence length is chosen as 200 according to the data statistics. For demographics data, we consider the following features: patient gender, age and zip codes. For location-based public health features, we use in total 34 different features as described in Section \ref{sec:dataset}. The embedding dimension for both demographics features and location-based public health features are set as 2. For training all approaches, we use Adam~\cite{kingma2014adam} with the batch size of 256 and the learning rate of $1e\text{-}3$. We train all the models until convergence. We use the cross entropy loss. All models are trained on Ubuntu 16.04 with 128GB memory and a Nvidia Tesla P100 GPU. \\

% We implement all models with Pytorch. We split the whole dataset into $80\%$ training and $20\%$ testing. For the training set, we further sample $20\%$ for validation.  For all models, we use Adam optimizer \cite{kingma2014adam} with a batch size of 256. For all models, we use the cross entropy loss.  All models are trained on Ubuntu 16.04 with 128GB memory and a Nvidia Tesla P100 GPU. 

\noindent\textbf{\ours Setup}.
To embed patient EHR records data, we use 2 blocks of transformers, each block applies a multi-head self attention layer, a normalization layer, a feed forward layer and another normalization layer. 
For demographics and location-based feature embeddings, we both set 2 fully-connected layers to learn the latent representations. The number of inducing points is set as 200. The hidden dimension for fused vector is set to 16. For transformer, we set the depth as 2 and number of heads as 8.\\

\noindent\textbf{Baseline Setup}. 
We implement \RETAIN\footnote{\url{https://github.com/easyfan327/Pytorch-RETAIN}}, \Dipole\footnote{\url{https://github.com/yzhao062/PyHealth}}, \MiME\footnote{\url{https://github.com/mp2893/mime}}, \DE\footnote{\url{https://github.com/Kyushik/Predictive-Uncertainty-Estimation-using-Deep-Ensemble}}, \MC\footnote{\url{https://github.com/cpark321/uncertainty-deep-learning}}, and \DMEGP\footnote{\url{https://github.com/jik0730/Deep-Mixed-Effect-Model-using-Gaussian-Processes}} based on the available  code repositories. For \INPREM, we acquired the code from the authors. We implement \LSTM ourselves. For each of the baseline models, we perform grid search for the best hyperparameters. 

\begin{table*}[h!]
\centering
\caption{\textbf{Exp 1. Performance Comparison}. All model performance are based on features from all modalities. \ours has the best performance thanks to its  efficient data fusion module that captures individual-level features, and the uncertainty modeling that learns population-level correlation.
}
% \vspace{-0.1in}
\resizebox{\textwidth}{!}{
\begin{tabular}{lccc|ccc} 
\toprule 
& \multicolumn{3}{c|}{Nonalcoholic fatty liver disease (NASH)} & \multicolumn{3}{c}{Alzheimer's disease (AD)} \\
            &   F1 Score & Cohen's Kappa  & PR AUC  & F1 Score & Cohen's Kappa  & PR AUC \\
\midrule
\LSTM       & 0.309  $\pm$  0.031    & 0.209 $\pm$  0.069          & 0.358 $\pm$  0.095     & 0.678 $\pm$ 0.040   & 0.469 $\pm$  0.032  & 0.798    $\pm$ 0.015       \\
\MiME       & 0.458  $\pm$  0.050    & 0.335 $\pm$  0.068          & 0.483 $\pm$  0.047     & 0.760 $\pm$ 0.030   & 0.595 $\pm$  0.035  & 0.802    $\pm$ 0.013       \\
\Dipole     & 0.460  $\pm$  0.021    & 0.325 $\pm$  0.046          & 0.501 $\pm$  0.040    & 0.770 $\pm$ 0.012   & 0.615 $\pm$  0.014   & 0.827    $\pm$ 0.001       \\
\RETAIN     & 0.358  $\pm$  0.032    & 0.226 $\pm$  0.025          & 0.336 $\pm$  0.032     & 0.774 $\pm$ 0.014   & 0.616 $\pm$  0.009  & 0.822    $\pm$ 0.006      \\
\midrule
\DE & 0.505 $\pm$ 0.018& 0.410 $\pm$ 0.006&  0.561 $\pm$ 0.015& 0.746 $\pm$ 0.069& 0.584 $\pm$ 0.067& 0.824 $\pm$ 0.015
\\
\MC  & 0.490 $\pm$ 0.035& 0.397 $\pm$ 0.019& 0.546 $\pm$ 0.018& 0.774 $\pm$ 0.027& 0.621 $\pm$ 0.029 & 0.829 $\pm$ 0.010
\\
\DMEGP     & 0.472   $\pm$  0.018    & 0.368 $\pm$  0.021         & 0.452 $\pm$  0.031     & 0.792 $\pm$ 0.018    & 0.633 $\pm$ 0.026  & 0.780      $\pm$ 0.013       \\
\INPREM   &0.510 $\pm$ 0.033 & 0.363 $\pm$ 0.065 &  0.569 $\pm$ 0.014 & 0.824 $\pm$ 0.003 & 0.685 $\pm$ 0.004 & 0.843 $\pm$ 0.029 \\
\midrule
\texttt{UNITE}$_{d}$ & 0.494 $\pm$ 0.002   & 0.399 $\pm$ 0.009     & 0.517 $\pm$ 0.029   & 0.834 $\pm$ 0.002    & 0.703 $\pm$ 0.006   & 0.846 $\pm$ 0.017 \\  
\texttt{UNITE}$_{b}$  & 0.547 $\pm$ 0.009 & 0.462 $\pm$ 0.010  & 0.608 $\pm$ 0.004 & \bf 0.841 $\pm$ 0.002  & 0.716 $\pm$ 0.005 & \bf 0.869 $\pm$ 0.026 \\  \midrule
\ours  & \bf 0.572 $\pm$ 0.007 & \bf 0.486 $\pm$ 0.009 & \bf 0.609 $\pm$ 0.005 & \bf  0.841 $\pm$ 0.004   & \bf 0.725 $\pm$ 0.005 &  0.856 $\pm$ 0.020 	\\ 
% \% Improve & \bf \% $\uparrow$    & \bf \% $\uparrow$   &   \bf   \% $\uparrow$  & \bf   \% $\uparrow$  &  \bf   \% $\uparrow$  & \bf   \% $\uparrow$	\\
\bottomrule 
\end{tabular}}
\label{tab:all_result}
\end{table*}

% \noindent\textbf{Metrics}
\subsection{Metrics}
Since the datasets in the experiment are  imbalanced, we choose  metrics that are fit for measuring model performance under data imbalance setting.
\begin{enumerate}[leftmargin=*]
    \item \textbf{F1 Score}: $ = 2\cdot(\mathrm{Prec}\cdot\mathrm{Rec})/(\mathrm{Prec}+\mathrm{Rec})$, where $\mathrm{Prec}$ is precision and $\mathrm{Rec}$ is recall.
  \item \textbf{Cohen's Kappa ($\kappa$)}:  can be used to measure the performance of the different classifiers for the same dataset, especially useful for imbalanced dataset. It is computed as below.
    \begin{equation}
        \kappa =\frac{p_o-p_e}{1-p_e} 
    \end{equation}
    $p_o$ is the observed agreement, which is identical to accuracy. $p_e$ is the expected agreement, which is the probability of random seeing each category. 
    % \item \textbf{Area Under the Receiver Operating Characteristic Curve}~(ROC-AUC):
    % measures the probability that a classifier will rank a randomly chosen positive instance higher than a randomly chosen negative one.
        \item \textbf{Area Under the Precision-Recall Curve}~(PR-AUC):
    \begin{equation*}
        \text{PR-AUC} = \sum_{k = 1}^{n} \mathrm{Prec}(k) \Delta \mathrm{Rec}(k), 
    \end{equation*}
    where $k$ is the $k$-th precision and recall operating point.
\end{enumerate}

% \noindent (1) \underline{F1 Score}; and (2) \underline{Area Under the} \underline{ Precision-Recall Curve}~(PR-AUC): popular measures of a model's accuracy based on precision and recall, better for imbalanced setting; 
% (3) \underline{Cohen's Kappa ($\kappa$)}:  $\kappa=(p_o-p_e)/(1-p_e)$, where $p_o$ is the observed agreement (i.e., accuracy), and $p_e$ is the expected agreement, which is the probability of randomly seeing each category.

\subsection{Results} 

\subsection*{Exp 1. Performance Comparison}

We experimentally evaluate \ours against leading deep learning based risk prediction models, as well as the state-of-the-art probabilistic deep learning models with uncertainty estimation on two tasks: NASH detection and AD detection. NASH is more difficult to diagnose without biopsy even for experienced clinicians because many NASH patients are asymptomatic~\cite{Madrazo17}.  We choose the two disease detection tasks with different difficulty levels to better understand the benefit of \ours compared with other methods.
We report the results in Table~\ref{tab:all_result}. Results are averaged over 3 random runs. The best results are presented in bold figures. 

On both datasets, \ours significantly outperforms the best baseline method by up to $19\%$ in Cohen's Kappa. The performance gains of \ours are attributed to  the well-designed embedding structure and the efficient data fusion module to model the individual-level features, as well as the uncertainty modeling provided by GP, which enables the model to learn population-level correlation.\\

\noindent
\textbf{Significance Testing.} To validate our claim that \ours outperforms all the baseline methods across all the metrics (e.g. the difference between the result of \ours and the baseline is larger than zero). We conduct a t-test with significance level $\alpha$ as $1\%$. For each of the baseline, we first calculate the $t$ statistic of the differences between the baseline results with the \ours result for all metrics. With the $t$ statistic, we calculate the $p$-values for all baseline methods and report them as follows: (1) \LSTM: $p=0.0009$; (2) \MiME: $p=0.0004$;  (4) \Dipole: $p=0.0014$; (6) \RETAIN: $p=0.0063$; (3) \DE $p=0.0021$; (5) \MC: $p=0.0004$;  (7) \DMEGP: $p=0.0006$; (8) \INPREM: $p=0.0085$. All of them are below the significance level $1\%$. Hence, we reject the null hypothesis and accept the alternative hypothesis, e.g. true mean is greater than zero. This indicates \ours is significantly better than all baselines.\\

% run three replications of our experiments with the same setting. We calculate the confidence intervals for each of the case given by $\Bar{x}\pm t\frac{s}{\sqrt{n}}$. The results turn out that except for AD with PR AUC, for all other other cases and metrics, we have the lower bounds of the $90\%$ confidence interval larger than zero. Hence, overall, \ours outperform the best baselines. \cx{list results/p value for t test for \ours versus all baselines. }

Overall, deep learning models with uncertainty estimations perform better than  the models without uncertainty modeling since Bayesian modeling can  better capture individual patterns and thus provide more accurate predictions.  Compared with  existing uncertainty estimation models, \ours further outperforms all the baselines. With the benefit from a well-designed embedding structure in our deep kernel function, GP aims to give predictions and uncertainties by learning the global correlation among all the patients, while other uncertainty modeling methods do not capture the population-level relationships. Although \DMEGP also uses GP, it does not learn the population level uncertainty but models the global trend with a multilayer perceptron neural network.

Between the two tasks, \ours provides more performance improvement compared with baselines on  NASH detection,
which shows \ours benefits more for diseases that are more uncertain since  NASH is difficult to diagnose without biopsy. With the predictive variances, \ours and other Bayesian models are  more robust in terms of model mis-specification and out-of-distribution examples.

In addition, we can observe that even the variant \texttt{UNITE}$_d$ (i.e., it replaces the variational inference module with a fully connected classifier) can achieve comparable performance to baselines on the NASH dataset and even better on the AD dataset. In other words, without employing GP, only using the multi-sourced data is effective for risk prediction. To further improve the performance and provide the utility of the designed model, we investigate the uncertainty-based variant \texttt{UNITE}$_b$, which estimates uncertainty with Monte Carlo sampling to approximate Bayesian NNs, performs better than \texttt{UNITE}$_d$. This result demonstrates considering uncertainty in risk prediction is necessary. However, \ours achieves the best overall performance among all the baselines and its two variants, which conforms the reasonableness of the designed model and the effectiveness of GP for risk prediction.

\subsection*{Exp 2. Leveraging Uncertainty to Improve Prediction}

In clinical decision making where overconfident incorrect predictions can be harmful or offensive, it would be better to set the highly uncertain subjects aside for human experts to evaluate. Also, it is always better to continue improving the prediction score to reach some thresholds that are reliable enough for decision making.

To investigate whether leveraging \ours and the learned uncertainty score could potentially identify truly uncertain patients that are easily misdiagnosed and thus improve the prediction, we perform the analysis below. We remove the patients with high uncertainty and check the improvement of the performance without re-training the model. We remove the top $20\%$, $50\%$, and $80\%$ of samples based on their uncertainty scores respectively for comparison. The results are listed in Table~\ref{tab:rq3}. As expected, for NASH that is highly uncertain for diagnosis, leveraging uncertainty score and ruling out highly uncertain patients could significantly improve the all accuracy measures for the rest of the cohort.
For AD which has more distinguishing disease patterns, the performance could still be improved.

\begin{table}[tb]
\centering
\caption{\textbf{Exp 2. Leverage Uncertainty  to Improve Prediction.}  We remove the top $20\%$, $50\%$ and $80\%$ of patients based on their uncertainty scores, respectively. The model achieves higher score gradually, especially for NASH that is highly uncertain and thus difficult to detect.}
% \vspace{-0.1in}
\resizebox{0.48\textwidth}{!}{
\begin{tabular}{lccc} 
\toprule
% & \multicolumn{3}{c}{Nonalcoholic fatty liver disease (NASH)}  \\
  \textbf{NASH}                  &   F1 Score    & Cohen's Kappa   & PR AUC         \\
\midrule

0\%    & 0.572 $\pm$ 0.007 &	0.486 $\pm$ 0.009	& 0.609 $\pm$ 0.005	 \\          
20\%   & 0.592 $\pm$ 0.028 &	0.501 $\pm$ 0.037	& 0.587 $\pm$ 0.007	 \\         
50\%   & 0.608 $\pm$ 0.059 &	0.522 $\pm$ 0.076	& 0.602 $\pm$ 0.020	 \\          
80\%   &  \bf 0.621 $\pm$ 0.080 &  \bf	0.537 $\pm$ 0.100	& \bf 0.610 $\pm$ 0.012	 \\    
\midrule
% & \multicolumn{3}{c}{Alzheimer's disease (AD)}  \\
     \textbf{AD}               &   F1 Score    & Cohen's Kappa   & PR AUC         \\
\midrule

0\%    & 0.840 $\pm$ 0.004	 & 0.725 $\pm$ 0.005	& 0.856 $\pm$ 0.020 	 \\          
20\%   & 0.858 $\pm$ 0.032	 & 0.754 $\pm$ 0.060	& 0.890 $\pm$ 0.010 \\         
50\%   & 0.868 $\pm$ 0.051	 &  \bf 0.763 $\pm$ 0.079	& 0.894 $\pm$ 0.020	 \\          
80\%   &  \bf 0.880 $\pm$ 0.069	 &  \bf 0.763 $\pm$ 0.074	&  \bf0.905 $\pm$ 0.037	 \\    
\bottomrule
\end{tabular}}
\label{tab:rq3}
\end{table}

\subsection*{Exp 3. Effect of Multiple Data Sources}

\begin{figure}[h!]
  \includegraphics[width=.45\textwidth]{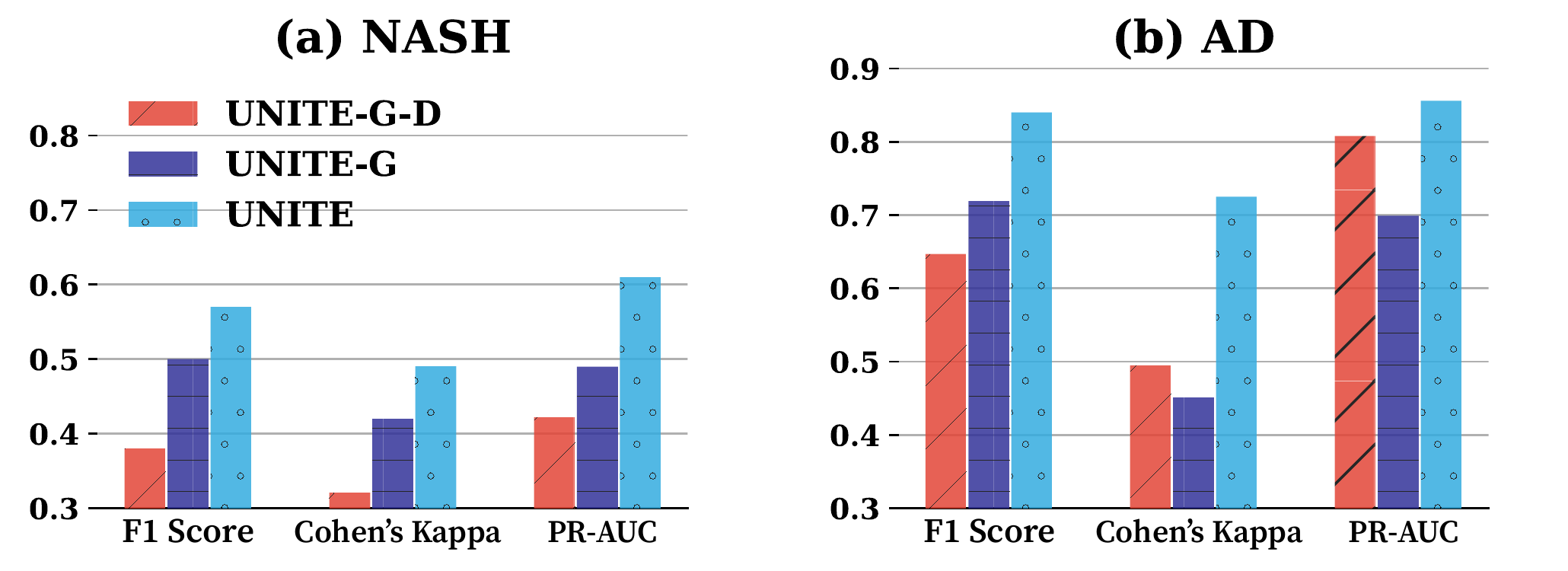} 
     \caption{\textbf{Exp 3. Effect of Multiple Data Sources.}}
    \label{fig:ablation_1}
    % \vspace{-0.1in}
\end{figure}

We conduct an ablation study to understand the contribution of each data sources in \ours. We remove/change the data modalities as below and compare the accuracy scores. The parameters in all the variants are determined with cross-validation, and the best performance is plotted in Fig.~\ref{fig:ablation_1}. 
We compare \ours with the reduced models below.
\begin{itemize}[leftmargin=*]
    \item \ours-G-D: without using location-based data $G$ and demographics $D$, only EHR sequences $E$ are used for training and inference.
    \item \ours-G:  without using location-based features, trained on patient EHR data $E$ and demographics $D$.
\end{itemize}

From Fig.~\ref{fig:ablation_1}, it can be observed when we solely use the patient EHR data for disease risk prediction, the performance drops largely especially for NASH which is hard to diagnose. It suggests the necessity of using all data including patient demographics and location-based features.

    \subsection*{Exp 4. Clustering Effect of \ours}
    
To investigate their effectiveness of learned covariance matrix, we compare two GP models, \ours and \DMEGP. We randomly sample an unseen dataset including 200 patients (100 positive and 100 negative) and expect that the models could perfectly cluster the population into 2 groups based on patients' similarities. The clustering effect of GP models is based on the learned covariance matrix $\Sigma = K_{VV}$, as specified in Eq.~\eqref{eq:rbf}. Element $\sigma_{i,j}$ in the $i$-th row and $j$-th column represents the correlation between sample $i$ and sample $j$. Larger value indicates higher correlation. The diagonal of $\sigma$ consists of the variance $\sigma_i^2$ of the $i$-th sample. 

\begin{figure}[h!]
  \centering
%   \vspace{-0.1in}
  \includegraphics[width=0.8\linewidth]{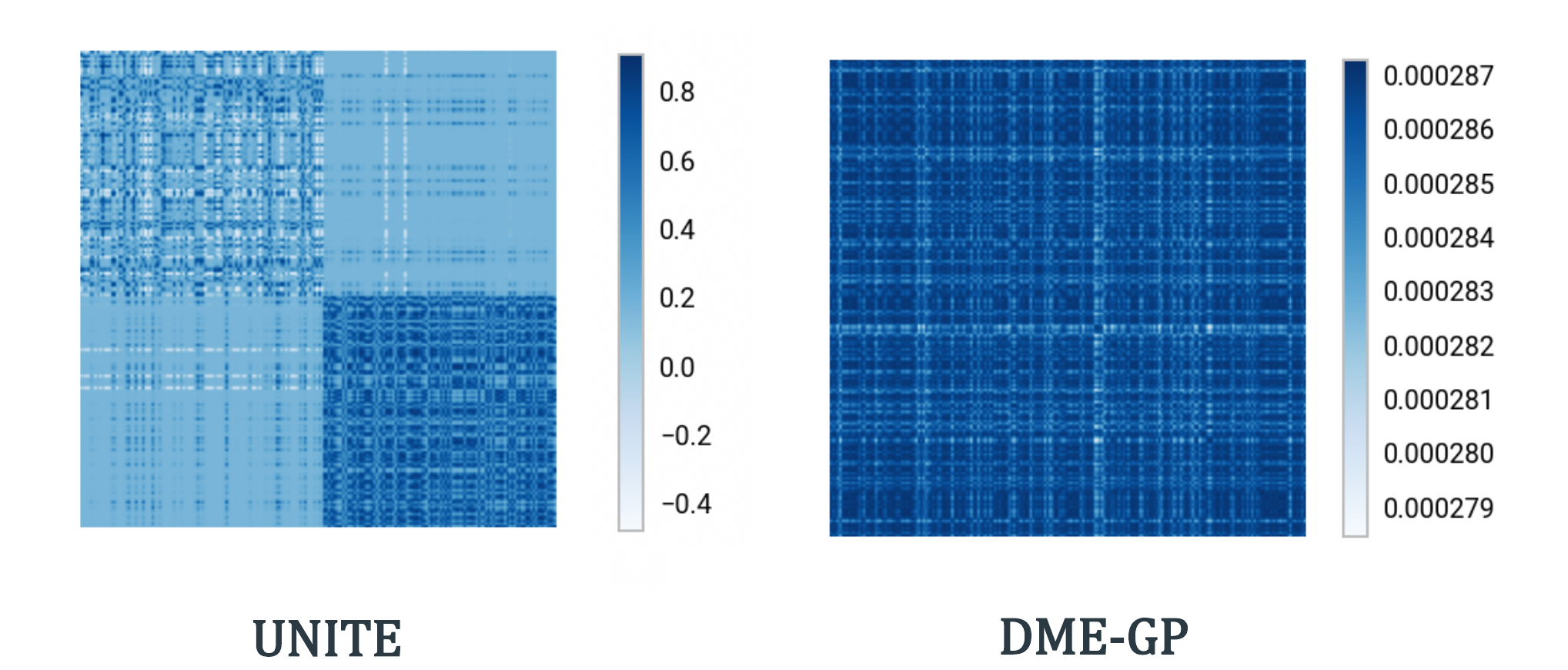}
%   \includegraphics[width=.5\linewidth]{figs/alz_2.pdf}
% \vspace{-0.1in}
  \caption{\textbf{Exp 4. Clustering Effect of \ours}. we leverage covariance matrix for similar patient retrieval. The covariance learned by \ours successfully clusters the patients, aligned with the ground-truth, while \DMEGP fails to model patient correlations.}
  \label{fig:rq3}
%   \vspace{-0.1in}
\end{figure}

To visualize the effect of learned correlations, we use spectral biclustering algorithm \cite{kluger2003spectral} to cluster the rows and columns of $\Sigma$ according to their values. As shown in Figure \ref{fig:rq3}, \ours succesfully clusters the population into 2 groups, which is aligned with the ground-truth. On the contrary, \DMEGP fails learn any meaningful correlations among patients. The reason is that \DMEGP applies both deep mean function and deep kernel function. The deep mean function is used to model the global trend. However, in this way, \DMEGP fails to model the population-level uncertainties. Our proposed \ours fully utilize the uncertainty modeling power given by the kernel functions.

\section{Conclusion}

In this paper, we propose \underline{UN}certa\underline{I}n\underline{T}y-based h\underline{E}alth risk prediction (\ours) model, that it built upon an adaptive multimodal deep kernel and a stochastic variational inference module, to provide accurate disease risk prediction and uncertainty estimation leveraging multi-sourced health data including public health data and patient EHR. Experiments on two disease detection tasks demonstrate the strong performance of \ours.

The \ours method can also be extended to other application domains where overconfident incorrect predictions can be harmful or offensive, such as fraud detection. Besides, there are several interesting future directions that need investigations. For example, prediction with uncertainty estimation for rare disease, in which highly imbalanced data and uncertain diagnosis introduce more difficulty. Another extension direction is uncertainty estimation for clinical notes comprehension where the we will need to solve new challenges from such a professional written unstructured data.

\section*{Acknowledgments}

This work was in part supported by the National Science Foundation award SCH-2014438, IIS-1418511, CCF-1533768, IIS-1838042, the National Institute of Health award NIH R01 1R01NS107291-01 and R56HL138415.

\bibliography{WWW-sample-sigconf}
\bibliographystyle{ACM-Reference-Format}

\end{document}